\documentclass{article}
\usepackage{spconf,amsmath,graphicx}

\title{Integrating discrete and neural features via mixed-feature trans-dimensional random field language models}
\name{Silin Gao$^1$, Zhijian Ou$^{1\dagger}$, Wei Yang$^2$, Huifang Xu$^3$\thanks{This work is supported by NSFC (No.61976122),  Key Technology Develop and Research Project of SGCC (No.5400-201953257A-0-0-00). $\dagger$ Corresponding author.}}
\address{$^1$Speech Processing and Machine Intelligence (SPMI) Lab, Tsinghua University, Beijing, China. \\
	$^2$State Grid Customer Service Center, $^3$China Electric Power Research Institute\\
gsl16@mails.tsinghua.edu.cn, ozj@tsinghua.edu.cn
}
\usepackage{scalerel,stackengine}
\stackMath
\newcommand\reallywidehat[1]{%
	\savestack{\tmpbox}{\stretchto{%
			\scaleto{%
				\scalerel*[\widthof{\ensuremath{#1}}]{\kern-.6pt\bigwedge\kern-.6pt}%
				{\rule[-\textheight/2]{1ex}{\textheight}}
			}{\textheight}%
		}{0.5ex}}%
	\stackon[1pt]{#1}{\tmpbox}%
}

\begin{document}
\ninept
\maketitle
\begin{abstract}
There has been a long recognition that discrete features (n-gram features) and neural network based features have complementary strengths for language models (LMs).
Improved performance can be obtained by model interpolation, which is, however, a sub-optimal two-step integration of discrete and neural features.
The trans-dimensional random field (TRF) framework has the potential advantage of being able
to flexibly integrate a richer set of features.
However, either discrete or neural features are used alone in previous TRF LMs.
This paper develops a mixed-feature TRF LM and demonstrates its advantage in integrating discrete and neural features.
Various LMs are trained over PTB and Google one-billion-word datasets, and evaluated in N-best list rescoring experiments for speech recognition.
Among all single LMs (i.e. without model interpolation), the mixed-feature TRF LMs perform the best, improving over both discrete TRF LMs and neural TRF LMs alone, and also being significantly better than LSTM LMs. 
Compared to interpolating two separately trained models with discrete and neural features respectively, 
the performance of mixed-feature TRF LMs matches the best interpolated model, and with simplified one-step training process and reduced training time.

\end{abstract}
\section{Introduction}
\label{sec:intro}

Language modeling (LM) involves determining the joint probability of words in a sentence, and is a crucial component in many applications such as speech recognition and machine translation. 
The directed graphical modeling approach is popular, decomposing the joint probability into a product of conditional probabilities. Examples include backoff n-gram LMs, particularly Kneser-Ney (KN) smoothed n-gram LMs \cite{chen1999empirical}, and neural network (NN) LMs \cite{schwenk2007continuous, mikolov2011extensions}, particularly Long Short Term Memory (LSTM) LMs \cite{sundermeyer2012lstm}.
Alternatively, in the undirected graphical modeling approach, trans-dimensional random field (TRF) LM has been developed \cite{wang2015trans,wang2017learning,wang2017language,wang2018learning,wang2018improved},
where sentences are modeled as a collection of random fields on subspaces of different dimensions/lengths
and the joint probability is defined in terms of potential functions.
Either linear potentials with discrete features (such as word n-gram features) \cite{wang2015trans,wang2017learning} or nonlinear potentials with neural network (e.g. LSTM) based features \cite{wang2017language,wang2018learning,wang2018improved} are used alone.
It has been shown that neural TRF LMs perform as good as LSTM LMs and are computationally more efficient (5x $\sim$ 114x faster) in inference (i.e. computing sentence probability).
The speed up factor mainly depends on the size of vocabulary (the larger the vocabulary size, the slower the inference by LSTM LMs), since LSTM LMs basically suffer from the high computational cost of the Softmax layer.

Generally, LMs with neural features (e.g. LSTM LMs, neural TRF LMs) outperform LMs with discrete features (e.g. KN LMs, discrete TRF LMs), but interpolation between them usually gives further improvement.
This suggests that discrete and neural features have complementary strengths.
Presumably, the n-gram features mainly capture local lower-order interactions between words, while the neural features particularly defined by LSTMs can learn higher-order interactions.
Additionally, by embedding words into continuous vector spaces, neural LMs are good at learning smoothed regularities, while discrete LMs may be better suited to handling symbolic knowledges or idiosyncrasies in human language, as noted in \cite{ostendorf2016continuous}.
Currently, model interpolation (linear or log-linear \cite{chen2019exploiting,wang2016model}) is often a second step, after the discrete and neural models are separately trained beforehand.
The interpolation weights are ad-hoc fixed or estimated over held-out data (different from the training data in the first step).
This two-step integration is sub-optimal.

In this paper, we propose a new principled approach to integrating discrete and neural features in LMs, based on the capability of the TRF modeling in flexibly integrating rich features.
Basically, with TRF modeling, one is free to define the potential function in any sensible way with much flexibility. It is straightforward to define a mixed-feature TRF LM, in which the potential function is a sum of a linear potential using discrete features and a nonlinear potential using neural features.
The new mixed-feature TRF LM can then be trained by applying the dynamic noise-contrastive estimation (DNCE) method \cite{wang2018improved}.
To the best of our knowledge, mixed-feature TRF LMs represent the first single LM model that incorporates both discrete and neural features without relying on a second-step interpolation.
Apart from naturally integrating discrete and neural features, another bonus from using mixed-feature TRF LMs is that, as shown in our experiments, we achieve faster training convergence and shorter training time, when compared to training neural TRF LMs alone. 
Notably, the log-likelihood of the training data with respect to (w.r.t.) the parameters of discrete features is concave. This helps to reduce the non-convexity of the optimization problem for maximum likelihood training.
Also, after incorporating the linear potential, the nonlinear potential only needs to capture the residual interactions between words. This may also explain the faster training convergence of mixed-feature TRF models.

Two sets of experiments over two training datasets of different scales are conducted to evaluate different LMs, by applying different LMs in N-best list rescoring  for speech recognition.
The discrete features used in discrete TRF LMs are word and class n-gram features as in \cite{wang2017language}.
For neural TRF LMs, we follow the same neural network architecture in \cite{wang2018improved} to define the potential function.

In the first set of experiment, various LMs are trained on Wall Street Journal (WSJ) portion of Penn Treebank (PTB) English dataset and then used to rescore the 1000-best list generated from the WSJ'92 test set, with similar experimental setup as in \cite{wang2018improved}.
Among all single LMs (i.e. without model interpolation), the mixed-feature TRF LMs perform the best, improving over both discrete TRF LMs and neural TRF LMs alone, and also being significantly better than LSTM LMs. 
This clearly demonstrates the benefit from integrating discrete and neural feature in one LM.
When comparing with the four interpolated model that result from combining LMs with neural features (LSTM LMs, neural TRF LMs) and LMs with discrete features (KN LMs, discrete TRF LMs), the performance of mixed-feature TRF LMs matches the best interpolated model, and with simplified one-step training process and reduced training time.
The results from the second set of experiment on the Google one-billion-word dataset \cite{chelba2013one} are similar.


\section{Related Work}
\label{sec:related}

Discrete features in language modeling mainly refer to word n-gram features, or more broadly, various types of linguistic indicator features such as word classes \cite{chen2009shrinking}, grammatical features \cite{amaya2001improvement}, and so on.
KN LMs, as classic LMs with discrete features, have been the state-of-the-art LMs during the last several decades \cite{chen1999empirical}, until recently the emergence of LMs with neural features (LSTM LMs, neural TRF LMs) which have been shown to significantly beat KN LMs \cite{mikolov2011extensions,wang2017language,wang2018improved}.

There has been a long recognition that discrete and neural features have complementary strengths for language modeling.
Improved performance can be obtained by interpolating KN LMs and LMs with neural features \cite{wang2016model,oparin2012performance,beck2019lstm}, which is, however, a sub-optimal two-step integration of discrete and neural features.
The (conditional) maximum entropy (ME) framework \cite{chen2009shrinking,khudanpur2000maximum} once appeared to be suited to integrating different types of features but with limited success, since these models basically suffer from the expensive computation of local normalization factors. This computational bottleneck hinders their use in practice.

The TRF framework eliminate local normalization from the root and has the potential advantage of being able
to flexibly integrate a richer set of features.
However, either discrete features (word and class n-gram features) \cite{wang2015trans,wang2017learning} or neural network (LSTM) based features \cite{wang2017language,wang2018learning,wang2018improved} are used alone in previous TRF LMs.
This paper develops mixed-feature TRF LMs and demonstrates the advantage of integrating discrete and neural features in the TRF framework.

\section{Mixed-feature TRF LMs}
\label{sec:mixTRF}

Throughout, we denote by $x^l=(x_1, \cdots, x_l)$ a sentence (i.e. word sequence) of length $l$, ranging from $1$ to $L$.
In the TRF framework \cite{wang2017learning,wang2018improved}, we assume that sentences of length $l$ are distributed from an exponential family model:
\begin{equation} \label{eq:mix-TRF}
p_{m}(l,x^{l};\lambda,\theta,\zeta)=\pi_{l}e^{\lambda^T  f(x^{l})+\phi(x^{l};\theta)-\zeta_{l}}
\end{equation}
where
$\pi_{l}$ is the empirical prior probability for length $l$.
Denote by $\zeta_{l}$ the logarithmic normalizing term for length $l$, and $\zeta = (\zeta_{1}, \zeta_{2}, ..., \zeta_{L})$. 

A remarkable flexibility of the TRF framework is that one is free to define the potential function in any sensible way. Eq. (\ref{eq:mix-TRF}) defines a mixed-feature TRF LM, in which the potential function is a sum of a linear potential $\lambda^T  f(x^{l})$ using discrete features and a nonlinear potential $\phi(x^{l};\theta)$ using neural features.
$f(x^l)$ is a vector of discrete features, which are computable functions of $x^l$ such as calculated by n-gram counts (explained in detail in \cite{wang2017learning}), $\lambda$ is the corresponding parameter vector.
We use a multi-layer bidirectional LSTM network to define $\phi(x^{l};\theta)$, with parameters $\theta$, as introduced in \cite{wang2018improved}. 
Let $e_{i}$ be the embedding vector for word $x_i$ in sentence $x^{l}$, and $h_{f,i},h_{b,i}$ the hidden vectors from the final forward and backward LSTM layers at position $i$, respectively. 
Then the nonlinear potential $\phi(x^l;\theta)$ is defined as follows:
\begin{equation}\label{eq:phi}
\phi(x^l;\theta) = \sum_{i=1}^{l-1} h_{f,i}^T e_{i+1} + \sum_{i=2}^{l} h_{b,i}^T e_{i-1}
\end{equation}

By referring to $f(x)$ as a feature vector, the terminology of discrete features is clear. 
We use the terminology of neural features in the sense that $\phi(x^l;\theta)$ is calculated from neural network based quantities from a sentence, as can be seen from Eq.(\ref{eq:phi}).

\begin{table*}[h]
	\centering
	\begin{tabular}{l|c|c|c|l|l}
		\hline
		Model                 &   PPL         &   WER (\%)  &   \#param (M)  & Training time       & Inference time \\
		\hline
		KN5                   &   141.2       &   8.78      &   2.3          & 22 seconds (1 CPU)  &  0.06 seconds (1 CPU) \\
		LSTM-2$\times$1500    &   78.7        &   7.36      &   66.0         & 23.6 hours (1 GPU)  &  9.09 seconds (1 GPU) \\
		\hline
		Discrete TRF in \cite{wang2017learning} & $\geq$130   &   7.90      &   6.4          & 24 hours (8 CPUs)    &  0.16 seconds (1 CPU) \\
		Neural TRF in \cite{wang2017language}   & $\geq$37    &   7.60      &   4.0          & 72 hours (1 GPU)       &  0.40 seconds (1 GPU) \\
		\hline
		Discrete TRF basic     & $\sim$128   &   8.37      &   2.3           & 7.28 hours (8 CPUs and 1 GPU)     &  0.11 seconds (1 CPU)\\
		Discrete TRF full      & $\sim$111   &   7.88      &   7.3           & 15.9 hours (8 CPUs and 1 GPU)     &  0.14 seconds (1 CPU)\\
		Neural TRF             & $\sim$75    &   7.34      &   2.6           & 22.1 hours (1 GPU)       &  0.08 seconds (1 GPU) \\
		Mixed TRF             & $\sim$69    &   7.17      &   4.9           & 18.2 hours (8 CPUs and 1 GPU)  &  0.12 seconds (1 CPU and 1 GPU)\\
		\hline
	\end{tabular}
	\vspace{-5pt}
	\caption{Speech recognition results of various LMs, trained on PTB dataset, by 1000-best list rescoring.
		``PPL'' is the perplexity on PTB test set.
		``WER'' is the rescoring word error rate on WSJ'92 test data.
		``\#param'' is the number of parameters (in millions).
		``Training time'' is the total time for training a LM. 
	``Inference time'' is the average time of rescoring the 1000-best list for each utterance.
CPUs denotes CPU-cores.}
	\label{tab:ptb1}
	\vspace{-10pt}
\end{table*}

\section{Model training with DNCE}
\label{sec:dncetraining}

\begin{figure}[t]
	\centering
	\includegraphics[width=1.0\linewidth]{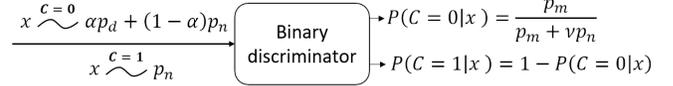}
	\vspace{-20pt}
	\caption{The illustration of the DNCE algorithm.
		The two differences between NCE and DNCE are 1)
		a dynamic noise distribution $p_n$ is introduced and jointly trained with the model distribution $p_m$; 2) DNCE discriminates between samples from the noise distribution and samples from the interpolation of the data distribution $p_d$ and the noise distribution $p_n$.
		Here a sample is a sentence, denoted by $x = (l,x^l)$.
	}
	\label{fig:dnce}
	\vspace{-10pt}
\end{figure}

The new mixed-feature TRF LM can be trained by applying the dynamic noise-contrastive estimation (DNCE) method \cite{wang2018improved}, as illustrated in Fig. \ref{fig:dnce}.
There are three distributions involved in DNCE -- the true but unknown data distribution $p_d(l,x^l)$, the model distribution $p_{m}(l,x^{l};\xi)$ with $\xi=(\lambda,\theta,\zeta)$ as in Eq.(\ref{eq:mix-TRF}), and a dynamic noise distribution $p_n(l,x^l;\mu)$. The noise distribution is defined by using a LSTM language model $p_{LSTM}(x^{l};\mu)$ with parameters $\mu$,  
namely $p_{n}(l,x^{l};\mu)=\pi_{l}p_{LSTM}(x^{l};\mu)$, where $\pi_{l}$ is the empirical prior length probability.

DNCE learns by discriminating sentences $(l,x^l)$ from two classes - $C=1$: the noise distribution, and $C=0$: the interpolated data distribution $p_d$ and noise distribution $p_n$, $q(l,x^{l};\mu)=\alpha p_{d}(l,x^{l})+(1-\alpha)p_{n}(l,x^{l};\mu)$, where $0 < \alpha < 1$ is the interpolating factor.
Assume that the ratio between the prior probabilities $P(C=1) / P(C=0) = \nu$, and the class-conditional probabilities for $C=1$ and $C=0$ are modeled by $p_n$ and $p_m$ respectively.
Then the posterior probabilities can be obtained as
$P(C=0|l,x^{l};\xi,\mu)=p_{m}/(p_{m}+\nu p_{n})$ and $P(C=1|l,x^{l};\xi,\mu)=1-P(C=0|l,x^{l};\xi,\mu)$.

DNCE estimates the model distribution by maximizing the following conditional log-likelihood:
\begin{displaymath}
\begin{split}
J(\xi)&=\sum_{l=1}^{L}\sum_{x^{l}}q(l,x^{l};\mu)logP(C=0|l,x^{l};\xi,\mu)\\
&+\nu\sum_{l=1}^{L}\sum_{x^{l}}p_{n}(l,x^{l};\mu)logP(C=1|l,x^{l};\xi,\mu)
\end{split}
\end{displaymath}
which can be solved by applying minibatch-based stochastic gradient descent (SGD).
At each iteration, a set of data sentences, denoted by $D$, is sampled from $p_d$, with the number of sentences in $D$ denoted as $|D|$ .
Additionally, two sets of noise sentences are drawn from the noise distribution $p_n$, denoted by $B_1$ and $B_2$,
whose sizes satisfy $|B_1| = \frac{1-\alpha}{\alpha} |D|$ and $|B_2| = \frac{\nu}{\alpha} |D|$ respectively.
As a result, the union of $D$ and $B_1$ can be viewed as samples drawn from the interpolated distribution $q(l,x^l;\mu)$.
We apply Adam \cite{kingma2014adam} to update the parameters $\xi$ of the model, with the following stochastic gradients:
\begin{displaymath}
\begin{split}
\reallywidehat{\frac{\partial J(\xi)}{\partial\xi}} = &\frac{\alpha}{|D|}\sum_{(l,x^{l})\in D\cup B_{1}}P(C=1|l,x^{l};\xi,\mu)g(l,x^{l};\xi) \\
&-\frac{\alpha}{|D|}\sum_{(l,x^{l})\in B_{2}}P(C=0|l,x^{l};\xi,\mu)g(l,x^{l};\xi)
\end{split}
\end{displaymath}
where $g(l,x^{l};\xi)$ denotes the gradient of the potential function w.r.t. $\xi=(\lambda,\theta,\zeta)$.
The three gradient components are $f(x^{l})$, $\partial\phi(x^{l};\theta)/\partial\theta$, and $-\left(\delta(l=1),...,\delta(l=L)\right)$, w.r.t. $\lambda$, $\theta$ and $\zeta$, respectively.
The gradient of $\phi$ w.r.t. $\theta$ can be calculated via back proporgation through the LSTM network, and $\delta(l=k)$ equals to $1$ if $l=k$ and $0$ otherwise.

DNCE estimates the noise distribution by minimizing the KL divergence, $KL(p_{d}||p_{n})$, between the data distribution and the noise distribution, with the following stochastic gradients:
\begin{displaymath}
\reallywidehat{\frac{\partial KL(p_{d}||p_{n})}{\partial\mu}}
=-\frac{1}{|D|}\sum_{(l,x^{l})\in D}\frac{\partial}{\partial\mu}logp_{n}(l,x^{l};\mu)
\end{displaymath}

To sum up, the DNCE algorithm, originally developed in \cite{wang2018improved} for training neural TRF LMs, can be easily extended to train mixture-feature TRF LMs, including discrete TRF LMs which is a special case of mixture-feature TRF LMs.

\section{Experiments}
\label{sec:experiment}

Two sets of experiments over two training datasets of different scales (PTB, Google one-billion-word) are conducted to evaluate different LMs, by applying different LMs in n-best list rescoring for speech recognition over WSJ’92 test data.
The CPUs used in the following experiments are Intel Xeon E5 (2.00 GHz) and the GPUs are NVIDIA GeForce GTX 1080Ti.
Most experiment settings follows \cite{wang2018improved}, unless stated otherwise.

\begin{table}[t]
	\centering
	\begin{tabular}{l|c}
		\hline
		Model                                  & WER (\%) \\
		\hline
		Mixed TRF                              & 7.17 \\
		\hline
		LSTM-2$\times$1500 + KN5               & 7.47 \\
		Neural TRF + KN5                       & 7.30 \\
		LSTM-2$\times$1500 + Discrete TRF basic     & 7.15 \\
		Neural TRF + Discrete TRF basic           & 7.17 \\
		\hline
		LSTM-2$\times$1500 + Neural TRF        & 7.01 \\
		LSTM-2$\times$1500 + Neural TRF + KN5  & 6.89 \\
		LSTM-2$\times$1500 + Mixed TRF           & 6.83 \\
		LSTM-2$\times$1500 + Mixed TRF + KN5     & 6.82 \\
		\hline
	\end{tabular}
	\vspace{-5pt}
	\caption{More results (continuing from Table \ref{tab:ptb1}) of various interpolated LMs, trained on PTB and test over WSJ'92.
	``+'' denotes the log-linear interpolation with equal weights.}
	\label{tab:ptb2}
	\vspace{-10pt}
\end{table}

\begin{figure}[t]
	\centering
	\includegraphics[width=0.85\linewidth]{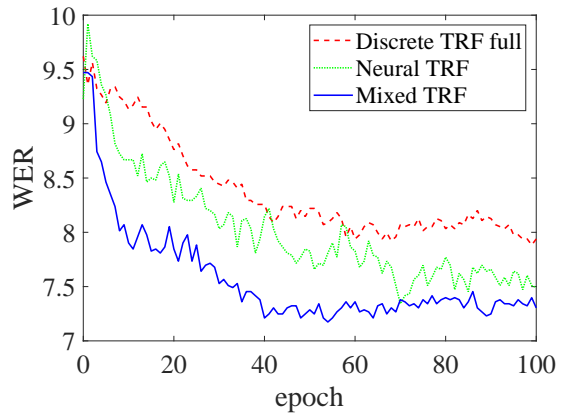}
	\vspace{-10pt}
	\caption{To show the training speed, the WER curves of the three TRF LMs during the first 100 training epochs are plotted.}
	\label{fig:wer}
	\vspace{-15pt}
\end{figure}

\begin{table*}[h]
	\centering
	\begin{tabular}{l|c|c|c|l|l}
		\hline
		Model                 &          PPL          &   WER (\%)  &   \#param (M)  & Training time              & Inference time \\
		\hline
		KN5                   &          94.5         &   6.13      &   133          & 2.48 hours (1 CPU)          &  0.491 seconds (1 CPU) \\
		LSTM-2$\times$1024    &          72.7         &   5.55      &   191          & 144 hours (2 GPUs)            &  0.909 seconds (2 GPUs) \\
		Neural TRF in \cite{wang2018improved}&$\sim$72&   5.47      &   114          & 336 hours (2 GPUs)           &  0.017 seconds (2 GPUs) \\
		\hline
		Discrete TRF basic    &        $\sim$86       &   6.04      &   102          & 131 hours (8 CPUs and 2 GPUs)  &  0.022 seconds (1 CPU) \\
		Mixed TRF             &        $\sim$68       &   5.28      &   216          & 297 hours (8 CPUs and 2 GPUs) &  0.024 seconds (1 CPU and 2 GPUs)\\
		\hline
	\end{tabular}
	\vspace{-5pt}
	\caption{Speech recognition results of various LMs, trained on Google one-billion-word dataset, by rescoring WSJ'92.
The columns have the same meaning as in Table \ref{tab:ptb1}.
}
	\label{tab:google1}
	\vspace{-15pt}
\end{table*}

\subsection{PTB dataset}
\label{ssec:exp-ptb}

In this experiment, the Wall Street Journal (WSJ) portion of Penn Treebank (PTB) \cite{marcus1993building} dataset is used, which contains about 1 million words. We split the data sections by 21:2:2 for training, development and test, and limit the vocabulary size to 10 K, including the special token ``$\langle$unk$\rangle$''.
Various LMs trained on PTB training and development sets are applied to rescore the 1000-best lists from recognizing WSJ'92 test data (330 utterances).

\textbf{Model descriptions.}
In Table \ref{tab:ptb1}, ``KN5'' denotes the Kneser-Ney smoothed 5-gram LM, and ``LSTM-2$\times$1500'' denotes the LSTM LM with 2 hidden layers and 1500 hidden units per layer, trained with the standard softmax output layer.
Two discrete TRF LMs are trained with DNCE, denoted as ``Discrete TRF basic'' and ``Discrete TRF full'', which use 5-order ``w+c'' and  ``w+c+ws+cs'' features respectively, by using the feature definition in \cite{wang2017learning}.
``w'' denotes word features.
``c'' denotes class features. 
Each word is deterministically assigned to a single class, by running the word clustering algorithm proposed in \cite{martin1998algorithms} on the training dataset.
``s'' denotes skipping features.
``Neural TRF'' is structured the same as in \cite{wang2018improved} but trained with different hyperparameters, so becomes slightly better than reported in \cite{wang2018improved}.
``Mixed TRF'' uses the same discrete features as in ``Discrete TRF basic'' and a bidirectional LSTM with 1 hidden layer and 200 hidden units to define the nonlinear potential, which is the same as in ``Neural TRF''.
For the noise distribution in DNCE, a LSTM LM with 1 hidden layer and 200 hidden units is used.

\textbf{Hyperparameter setup.}
The batch size $|D|=100$, and the initial learning rates for $\lambda$, $\theta$, $\zeta$ and $\mu$ are 0.003, 0.003, 0.01 and 1.0, respectively. 
We halve the learning rate of $\lambda$ and $\theta$ when the log-likelihood on the PTB development set does not increase significantly, and we stop the training when the learning rate of $\lambda$ and $\theta$ reduces to one tenth of the original. 
In the training of ``Discrete TRF basic'', ``Discrete TRF full'' and ``Neural TRF'', we set the interpolation factor $\alpha=0.25$ and the ratio factor $\nu=1$, while in the ``Mixed TRF'', we use $\alpha=0.2$.

\textbf{Results of single LMs.}
First, it can be seen from Table  \ref{tab:ptb1} that we can train discrete TRF LMs successfully via DNCE with much less training time than via AugSA \cite{wang2017learning}.
The performance is also improved when the parameter size is close. DNCE can handle not only neural features \cite{wang2018improved} but also discrete features, which is desirable for traing the new mixed-feature TRF LMs.

Second, among all single LMs (i.e. without model interpolation), ``Mixed TRF'' perform the best, improving over ``Discrete TRF'' (whether basic or full) and ``Neural TRF'', and also being significantly better than ``LSTM-2$\times$1500''.
The p-value from running matched-pairs test \cite{gillick1989some} between ``Mixed TRF'' and the suboptimal LM ``Neural TRF'' is less than 5\%.
This clearly demonstrates the benefit from integrating discrete and neural feature in one LM.
Moreover, although containing more parameters compared with ``Neural TRF'', ``Mixed TRF'' consumes less training time.
As shown in Fig. \ref{fig:wer}, ``Mixed TRF''  take as few epochs as ``Discrete TRF full'' to converge (but ``Mixed TRF'' converges to a lower WER). In contrast, ``Neural TRF''  learns much slower.
Notably, the log-likelihood of the training data w.r.t. the parameters of discrete features is concave, which lowers the non-convexity of the optimization problem for maximum likelihood training. 
Besides, the introduction of discrete features reduces the amount of patterns that the neural features need to capture. Presumably, the above two factors contribute to the fast training speed of mixed-feature TRF LMs.

\textbf{Results of interpolated LMs.}
We conduct log-linear interpolation on single LMs in Table \ref{tab:ptb1} with equal weights. Table 2 shows the rescoring WER results of these interpolated LMs. 
First, we interpolate a neural-feature LM (a LSTM LM or a neural TRF LM) with a discrete-feature LM (a KN LM or a discrete TRF LM). 
For comparison, we use ``Discrete TRF basic'' as the discrete TRF LM, since it uses the same set of discrete features as ``Mixed TRF''. 
There is a group of four such interpolated models.
The performance of mixed-feature TRF LMs match the best interpolated model in this group of models, and has the advantage of simplified one-step training process and reduced training time.
Notably, ``Mixed TRF'' alone significantly outperform ``LSTM-2$\times$1024 + KN5'', with the p-value less than 5\% from significance test.
Second, we examine more complex model interpolations.
It can be seen from Table \ref{tab:ptb2} that upgrading ``Neutal TRF'' to ``Mixed TRF'' as model components is beneficial in interpolations.
Numerically, ``LSTM-2$\times$1024 + Mixed TRF + KN5'' achieves the best performance of 6.82\% in WER.

\begin{table}[t]
	\centering
	\begin{tabular}{l|c}
		\hline
		Model                                  & WER (\%) \\
		\hline
		Mixed TRF                              & 5.28 \\
		\hline
		LSTM-2$\times$1024 + KN5               & 5.38 \\
		Neural TRF + KN5                       & 5.51 \\
		LSTM-2$\times$1024 + Discrete TRF basic      & 5.31 \\
		Neural TRF + Discrete TRF basic             & 5.27 \\
		\hline
		LSTM-2$\times$1024 + Neural TRF        & 5.25 \\
		LSTM-2$\times$1024 + Neural TRF + KN5  & 5.06 \\
		LSTM-2$\times$1024 + Mixed TRF           & 5.02 \\
		LSTM-2$\times$1024 + Mixed TRF + KN5     & 4.99 \\
		\hline
	\end{tabular}
	\vspace{-5pt}
	\caption{
More results (continuing from Table \ref{tab:google1}) of various interpolated LMs, trained on Google one-billion-word and test over WSJ'92.
``+'' denotes the log-linear interpolation with equal weights.
}
	\label{tab:google2}
	\vspace{-10pt}
\end{table}

\subsection{Google one-billion-word dataset}
\label{ssec:exp-google}

In this section, we examine the scalability of various LMs on Google one-billion-word dataset.
The training set contains about 0.8 billion words. We use a vocabulary of about 568 K words, after mapping the words whose counts less than 4 to ``$\langle$unk$\rangle$''.
Various LMs trained on the training set are used to rescore the WSJ'92 1000-best lists.
A LSTM LM which uses the embedding size of 256, 2 hidden layers and 1024 hidden units per layer (denoted by ``LSTM-2$\times$1024'') is trained using the adaptive softmax strategy proposed in \cite{grave2017efficient}.
The cutoff setting of ``00225'' is applied to the 5-gram features in ``KN5'', ``Discrete TRF basic'' and ``Mixed TRF''. 
The initial learning rates for $\lambda$, $\theta$, $\zeta$ and $\mu$ are 0.001, 0.001, 0.01 and 0.1, respectively. We halve the learning rate of $\lambda$ and $\theta$ per epoch. The final training epochs for ``Discrete TRF basic'' and ``Mixed TRF'' are 5 and 3. Besides, sentences longer than 60 words are omitted for the training of TRF LMs. The interpolation factor $\alpha=2/3$ and the ratio factor $\nu=4$ are the same as in \cite{wang2018improved}.

The experimental results of single LMs and interpolated LMs are shown in Table \ref{tab:google1} and \ref{tab:google2} respectively, which are similar to results over PTB.
Among all single LMs, ``Mixed TRF'' performs the best, takes shorter training time than ``Neural TRF'', and still matchs the best interpolation of combining a discrete LM and a neural LM.



\section{Conclusion}
\label{sec:conclusion}

This paper develops a mixed-feature TRF LM and demonstrates its advantage in integrating discrete and neural features.
Various LMs are trained over PTB and Google one-billion datasets, and evaluated in N-best list rescoring experiments for speech recognition.
Among all single LMs, the mixed-feature TRF LMs perform the best, improving over both discrete TRF LMs and neural TRF LMs alone, and also being significantly better than LSTM LMs. 
Compared to interpolating two separately trained models with discrete and neural features respectively, 
the performance of mixed-feature TRF LMs match the best interpolated model, and with simplified one-step training process and reduced training time.
Integrating richer features in the TRF framework and deploying the mixed-feature TRF LM in one-pass decoding are interesting future work.

\vfill\pagebreak

\bibliographystyle{IEEEbib}
\bibliography{refs}

\end{document}